\documentclass[10pt,twocolumn,letterpaper]{article}

\usepackage{cvpr}              
\definecolor{cvprblue}{rgb}{0.21,0.49,0.74}
\usepackage[pagebackref,breaklinks,colorlinks,allcolors=cvprblue]{hyperref}
\usepackage{multirow}
\usepackage{makecell}
\usepackage{xcolor}
\usepackage{graphicx}
\usepackage{tabularx}
\usepackage{array} 
\usepackage{colortbl}
\usepackage[accsupp]{axessibility}  
\def\paperID{6471} 
\def\confName{CVPR}
\def\confYear{2026}

\title{Leveraging Arbitrary Data Sources for AI-Generated Image Detection Without Sacrificing Generalization}

\author{
  Qinghui He\textsuperscript{\rm 1, 2}, ~
  Haifeng Zhang\textsuperscript{\rm 1, 2}, ~
  Xiuli Bi\textsuperscript{\rm 1, 3}, ~
  Bo Liu \thanks{~Corresponding Author}~~\textsuperscript{\rm , 1, 3}, ~
  Chi-Man Pun\textsuperscript{\rm 4}, ~
  Bin Xiao\textsuperscript{\rm 1, 5} \\
  \textsuperscript{\rm 1}Chongqing Key Laboratory of Image Cognition, Chongqing University of Posts and Telecommunications\\
  \textsuperscript{\rm 2}School of Computer Science and Technology ~~
  \textsuperscript{\rm 3}School of Artificial Intelligence ~~
  \textsuperscript{\rm 4}University of Macau\\
  \textsuperscript{\rm 5}Jinan Inspur Data Technology Co., Ltd., Jinan, China\\
  {\tt\small D250201011@stu.cqupt.edu.cn ~~~ boliu@cqupt.edu.cn}\\
}

\begin{document}
\maketitle
\begin{abstract}
The accelerating advancement of generative models has introduced new challenges for detecting AI-generated images, especially in real-world scenarios where novel generation techniques emerge rapidly. Existing learning paradigms are likely to make classifiers data-dependent, resulting in narrow decision margins and, consequently, limited generalization ability to unseen generative models. We observe that both real and generated images intend to form clustered low-dimensional manifolds within high-level feature spaces extracted by pre-trained visual encoders. Building on this observation, we propose a single-class attribution modeling framework that first amplifies the intrinsic differences between real and generated images by constructing a compact attribution space from any single-class training set, either composed of real images or generated ones, and then establishes a more stable decision boundary upon the enlarged separation. This process enhances class distinction and mitigates the reliance on generator-specific artifacts, thereby improving cross-model generalization. Extensive experiments show that our method generalizes well across various unseen generative models, outperforming existing detectors by as much as 7.21\% in accuracy and 7.20\% in cross-model generalization.
\end{abstract}    
\section{Introduction}
\label{sec:intro}

The rapid evolution of generative models, ranging from generative adversarial networks (GANs) to diffusion models (DMs), has significantly lowered the threshold for creating high-fidelity visual content than ever before, resulting in an increasingly blurred line between real and generated visual content. While enjoying the rich visual experience of generative models, users are also beginning to worry about potential abuse risks, including the spread of false information, political manipulation, and public trust crises. Therefore, there is an urgent need to develop efficient and reliable methods for detecting generated images to ensure the authenticity and reliability of the image information.

\begin{figure}[t]
    \centering
    \includegraphics[width=1\linewidth]{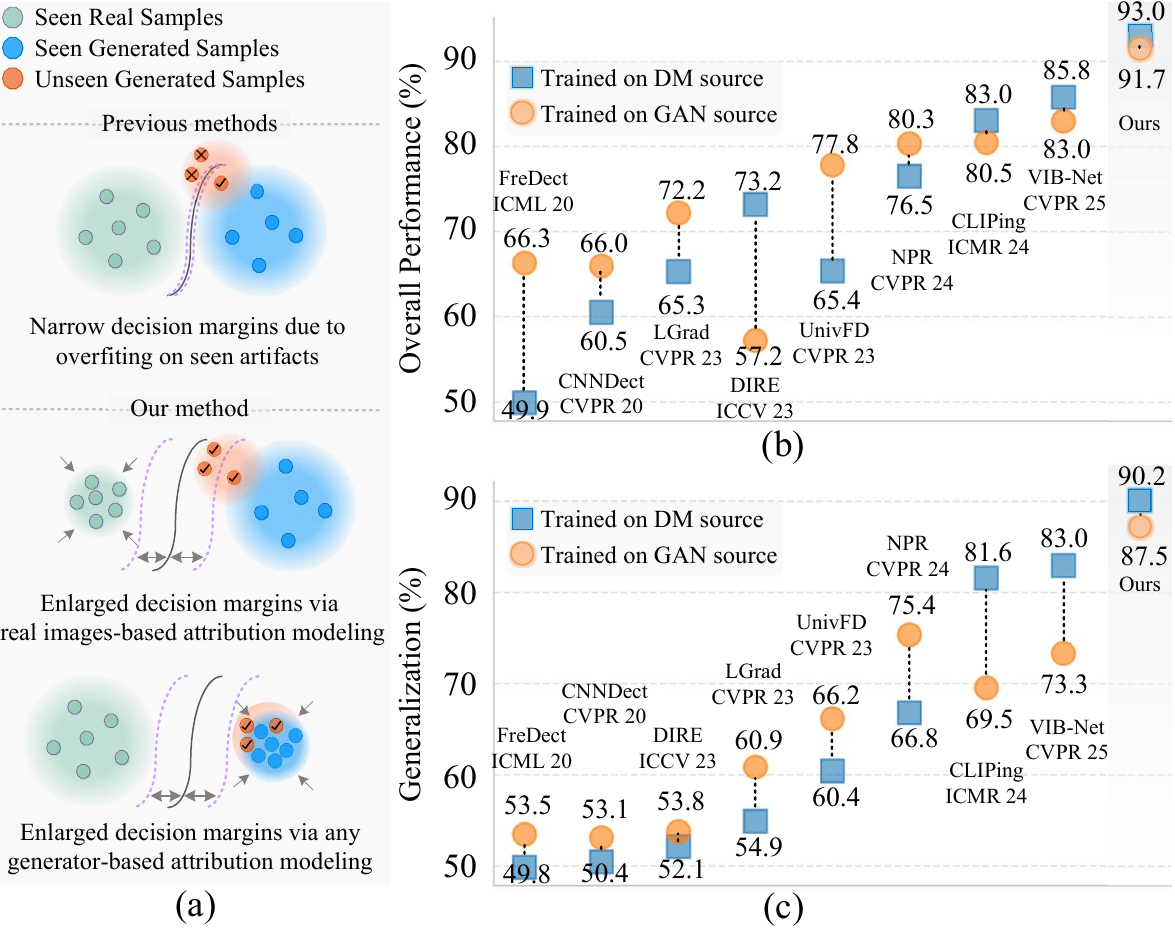}
    \caption{(a) Previous methods overfit to training distributions with small margins, while our attribution space yields larger margins and stronger cross-model generalization. (b) The overall performance of our method across the entire dataset compared to the SOTA methods. (c) The cross-model generalization performance of our method compared to the SOTA methods.}
    \label{fig1}
\end{figure}

Previous methods focus on learning subtle real–fake differences from large datasets~\cite{wang2020cnn,corvi2023detection} or on capturing anomalies in the frequency domain~\cite{frank2020leveraging,chandrasegaran2021closer,liu2022detecting}. Some works~\cite{ojha2023towards,khan2024clipping,cozzolino2024raising,zhang2025towards} rely on pre-trained models such as CLIP~\cite{radford2021learning} to extract semantic information, while others~\cite{wang2023dire,luo2024lare,cazenavette2024fakeinversion} use reconstruction mechanisms for forgery discrimination. However, these methods generally overfit to artifacts tied to specific generative architectures. Although effective under matched training and testing conditions, such methods exhibit limited cross-model generalization when applied to unseen generated samples, domain shifts, or new image styles. This reliance on diverse generated data severely constrains their practicality, especially in open-world detection scenarios where future techniques are unpredictable.

In this work, we explore a fundamentally different perspective on AI-generated image detection. Instead of directly modeling the diversity of generated artifacts left by generative models, we investigate whether it is possible to construct a general attribution space that effectively distinguishes real from generated images without overfitting to specific generated datasets. We observe that generated and real images tend to form their own clustered low-dimensional manifolds within high-level feature spaces extracted by pre-trained visual encoders. This insight motivates our attribution consistency paradigm, which leverages the intrinsic manifold of a single-class dataset to create a representation space where attribution deviations from out-of-distribution samples are naturally amplified. As shown in Fig.~\ref{fig1}(a), previous methods often suffer from narrow decision margins, leading to poor generalization to unseen generated samples. In contrast, our method first amplifies the intrinsic discrepancies between real and generated images by constructing a compact attribution space from a single-class dataset, either real or generated, and then establishes a more stable decision boundary based on this enlarged separation. Our method is highly flexible, as the attribution space can be constructed using images from a single generator or even real images.

Experimental results show that the proposed method achieves both strong detection capability and remarkable cross-model generalization, as illustrated in Fig.~\ref{fig1}(b) and Fig.~\ref{fig1}(c), demonstrating its robust detection performance across a wide range of generative models. Our contributions can be summarized as follows:
\begin{itemize}
    \item We break the traditional paradigm of explicitly modeling diverse generative artifacts by strategically inducing controlled manifold-aligned overfitting toward a stable single-class manifold, transforming overfitting from a limitation into a robust generalization mechanism. 
    \item By fitting any single-class distribution to construct the attribution space, our method amplifies the intrinsic separation between in- and out-of-distribution samples, yielding wider decision margins.
    \item Extensive experiments validate the effectiveness of our method, surpassing the current SOTA by 7.20\% in detection accuracy across a set of 15 generative models.
\end{itemize}
\section{Related Work}
\label{sec:formatting}

\subsection{Types of Generated Images}

Early GANs~\cite{brock2018large,zhu2017unpaired,karras2018progressive,karras2019style} achieved impressive image synthesis through adversarial training but were limited by data distribution and prone to mode collapse. Subsequently, autoregressive models~\cite{razavi2019generating,esser2021taming,ramesh2022hierarchical} expanded generative capability but suffered from low efficiency or limited resolution.
Recently, diffusion models~\cite{dhariwal2021diffusion,nichol2022glide,rombach2022high,gu2022vector} have achieved breakthroughs in fidelity, diversity, and stability, becoming the dominant generation paradigm.
Open-source models like Stable Diffusion and commercial tools such as Midjourney have further popularized high-quality AI-generated images among non-professional users.

\subsection{AI-Generated Image Detection}
\subsubsection{Image- and Frequency-based Methods.}
Early research focused on detecting images generated by GAN and its variants, and effectively detecting low-quality generated images by mining visible generation traces in generated images (such as incoherent textures, high-frequency anomalies, etc.).~\cite{wang2020cnn} demonstrated that a simple CNN classifier can effectively improve the generalization of unseen generated images by training images that have been post-processed, such as JPEG compression and blurring.~\cite{frank2020leveraging} started from the analysis of the frequency domain and revealed the recognizable traces that are commonly found in GAN-generated images.~\cite{chandrasegaran2021closer} demonstrated the significant difference in spectral distribution between GAN-generated images and real images by studying the upsampling mechanism of the generative model.~\cite{liu2022detecting} analyzed different types of images and found that real images are more consistent with generated images in the frequency domain.~\cite{corvi2023detection} found that the performance of detectors trained only on images generated by GANs drops sharply when generalized to images generated by diffusion models. This prompted researchers to explore detection frameworks with more generalization capabilities.

\subsubsection{Pre-trained Model-based Methods.}
To address cross-model generalization,~\cite{ojha2023towards} showed that feature spaces not explicitly trained for generated image detection (e.g., CLIP) generalize better to unseen models.~\cite{wu2025generalizable} combined CLIP's text-image contrast learning mechanism to improve adaptability, while~\cite{cozzolino2024raising} systematically studied the applicability of CLIP features in generated image detection task and proposed a detector requiring only limited data.~\cite{khan2024clipping} and~\cite{tan2025c2p} achieved strong generalization by adjusting CLIP text embeddings; ~\cite{zhang2025towards} achieved effective detection by filtering irrelevant information and retaining relevant information. However, these methods still tend to overfit to specific artifacts in the training set, resulting in poor generalization.~\cite{yan2024effort,he2026diversity} preserve a relatively high feature rank, thereby mitigating shortcut learning to some extent.

Since detectors have difficulty recognizing high-quality images generated by diffusion models, researchers have proposed detectors specifically for diffusion methods that take advantage of the inherent properties of diffusion models.~\cite{wang2023dire} showed that diffusion images can be reconstructed more accurately than natural images but require a costly multistep inversion.~\cite{luo2024lare} reduced this cost via single-step latent sampling. Similarly,~\cite{cazenavette2024fakeinversion} introduced inversion noise maps to enhance generalization.~\cite{ricker2024aeroblade} demonstrated that real images can be effectively distinguished from generated images using only some components inside the diffusion model.~\cite{chen2024drct, guillaro2025bias, gye2025reducing} focused on the classification of difficult samples by reconstructing the training dataset to enhance the generalization ability of detectors. Although these methods have the potential for generalization in certain scenarios, they usually require access to a specific generation pipeline and are limited to certain paradigms.
\section{Preliminaries}

\subsection{Motivation}
\begin{figure}[t]
    \centering
    \includegraphics[width=1\linewidth]{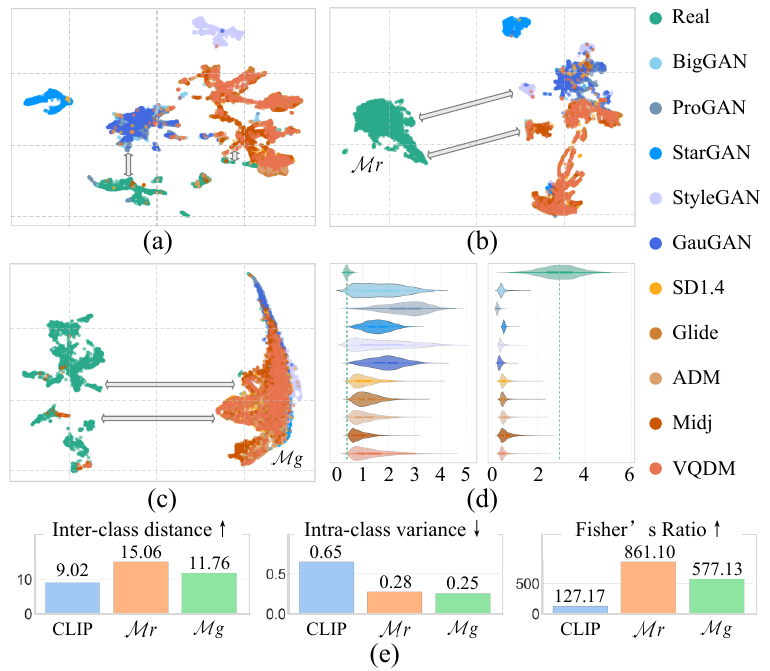}
    \caption{Visualization of CLIP features and attribution deviations. (a) UMAP visualization of CLIP features for real and generated images. (b) UMAP visualization of attribution deviations under $\mathcal{M}_r$. (c) UMAP visualization of attribution deviations under $\mathcal{M}_g$. (d) Violin plots of attribution deviations using $\mathcal{M}_r$ (left) and $\mathcal{M}_g$ (right). (e) Quantitative comparison of class separability using inter-class distance, intra-class variance, and Fisher’s Ratio.}  
    \label{fig2}
\end{figure}

We argue that existing AI-generated image detection methods often rely on artifacts or features associated with specific generative models, making them difficult to generalize across diverse generators and generation mechanisms. As shown in Fig.~\ref{fig2}(a), when features extracted by a pre-trained model (e.g., CLIP) are projected into a low-dimensional space, images generated by different models form distinct clusters, revealing the diversity of generative traces. However, the existing discriminative detectors tend to overfit to the generative cues seen during training and fail to recognize unseen ones~\cite{ojha2023towards}, especially when the architectures differ significantly.

From the dimensionality-reduction results, we further observe that different generative models introduce their own artifacts, leading to distinct clusters of generated images in the feature space. Meanwhile, real images are generally concentrated in one region, though a few samples show slight shifts. However, there remains a clear distributional gap between real and generated images. We marked these two regions with distance in Fig.~\ref{fig2}(a) to illustrate their separation in the feature space. This phenomenon suggests that real images share intrinsic semantic consistency, while generated images exhibit relatively uniform generative characteristics. Based on this observation, we hypothesize that real and generated images can each form a compact manifold representation through their shared internal features:
\begin{equation}
    f(x_i) \in \mathcal{M}_i, \quad \mathrm{dim}(\mathcal{M}_i) \ll D, i \in \{r, g\}.
\end{equation}
where $f(x)$ is the extracted feature, $\mathcal{M}_i$ is the manifold of class $i$, $D$ is the feature dimension, and $r,g$ denote real and generated classes, respectively.

To capture such intrinsic regularities, we propose a novel detection perspective based on attribution space modeling. The core idea is to construct an attribution space using only single-class images, allowing the model to learn the internal common features of that class. Features from another source are more likely to deviate from this manifold and produce measurable attribution shifts, formulated as: 
\begin{equation}
    f(x_j) \notin \mathcal{M}_i, \quad 
    \mathrm{Dis}(f(x_j), \mathcal{M}_i) \gg \mathrm{Dis}(f(x_i), \mathcal{M}_i).
\end{equation}
where $j \neq i$, $i,j \in \{r, g\}$, and $\mathrm{Dis}(\cdot,\cdot)$ measures the distance between a sample and the manifold. The inherent discrepancy between real and generated images leads them to form separable distributions in the attribution space, enabling detection without reliance on specific datasets.

\begin{figure*}[ht]
\centering
\includegraphics[width=0.95\textwidth]{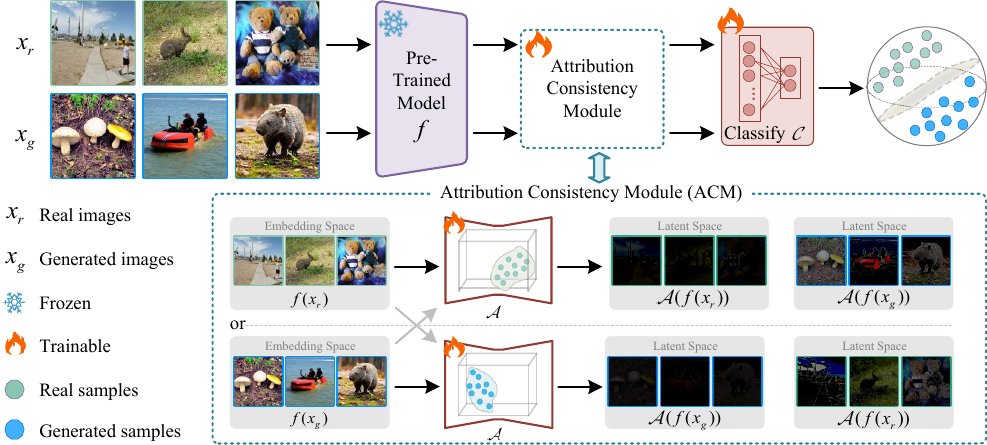}
\caption{Overview of the proposed framework. Given real images $x_r$ and generated images $x_g$, a frozen pre-trained model $f$ first extracts high-level semantic features. The proposed Attribution Consistency Module then constructs an attribution space $\mathcal{A}$ using samples from a single class (either real or generated) to learn class-consistent representations. Features from the opposite class that deviate from this learned manifold are treated as attribution shifts. A lightweight classifier $\mathcal{C}$ is finally applied to distinguish real and generated images based on their attribution-space representations.}
\label{fig3}
\end{figure*}

\subsection{Attribution Space Generalization Analysis}
Based on the above hypothesis, we evaluate whether the attribution spaces $\mathcal{M}_r$ and $\mathcal{M}_g$ can effectively capture intra-class consistency and enlarge the separation between real and generated images.
As shown in Fig.~\ref{fig2}(b)(c), when the attribution space is constructed using a single class, samples from that class form compact and coherent clusters, while samples from the opposite class are distributed farther apart, indicating clear inter-class separation.
Fig.~\ref{fig2}(d) presents the deviation distributions in $\mathcal{M}_r$ (left) and $\mathcal{M}_g$ (right), where the horizontal axis denotes the magnitude of deviation.
It can be observed that in both spaces, class-consistent samples maintain low deviation values, whereas samples from the other class exhibit significantly larger deviations.
For instance, in $\mathcal{M}_r$, real images show substantially smaller deviations than any generated images, while in $\mathcal{M}_g$, real images exhibit much larger deviation magnitudes than generated ones.
Furthermore, as shown in Fig.~\ref{fig2}(e), the margin metrics, including inter-class distance, intra-class variance, and Fisher’s ratio, demonstrate that compared to the original CLIP feature space, our attribution space maintains intra-class compactness while substantially increasing class separation, Fisher's ratio increasing by more than six times.
Overall, these results verify that the proposed attribution-space modeling effectively learns the common features within a single class and significantly enhances the discriminative boundary between real and fake images.

\section{Methodology}
\label{methodology}

This section elaborates on the proposed generated image detection framework, which is centered around the Attribution Consistency Module. We first outline the overall framework and then introduce the specific structure, training strategy, and discrimination mechanism of each component.

\subsection{Overall Framework}
Our method is built upon the key idea of learning class-consistent representations within a single-class attribution space, rather than relying on cross-class discriminative training. As illustrated in Fig.~\ref{fig3}, real and generated images ($x_r$ and $x_g$) are first encoded into a shared semantic feature space by a frozen pre-trained model $f$. The ACM then constructs an attribution space $\mathcal{A}$ from samples of a single class (either real or generated), capturing the intrinsic feature regularities of that class. When features from another source are projected into this space, they deviate from the learned manifold, resulting in measurable attribution discrepancies that reflect semantic inconsistency. A lightweight classifier $\mathcal{C}$ finally identifies image authenticity based on these attribution-space representations. This framework enables generalized and model-agnostic detection by transforming class-consistency learning into a robust discrimination mechanism.
During training, we employ an alternating optimization strategy between $\mathcal{A}$ and $\mathcal{C}$. Specifically, we first freeze $\mathcal{C}$ and refine $\mathcal{A}$ with the attribution loss $\mathcal{L}_\mathcal{A}$, then freeze $\mathcal{A}$ and update $\mathcal{C}$ using the classification loss $\mathcal{L}_\mathrm{classify}$. This iterative process allows the ACM to adaptively enhance feature separability and improve discrimination capability between real and generated images.

\begin{figure*}[ht]
\centering
\includegraphics[width=0.95\textwidth]{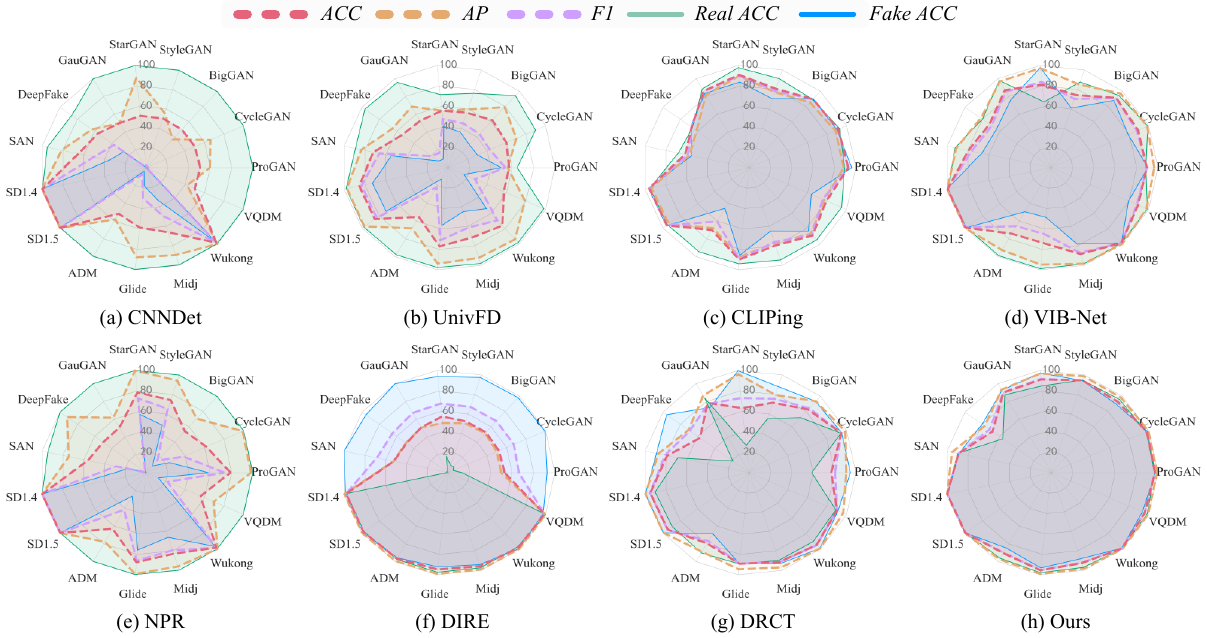}
\caption{Performance comparison of SOTA detectors (a-g) and our method (h) trained on SD1.4, evaluated on 15 generative models using \textit{ACC}, \textit{AP}, \textit{F1}, \textit{Real ACC}, and \textit{Fake ACC}. Our approach achieves balanced real-fake accuracy and superior cross-model generalization.}
\label{fig4}
\vspace{-5pt}
\end{figure*}

\subsection{Attribution Consistency Module}
We denote the training dataset as $D_s=\{(x^i,y^i)\}_{i=1}^N$, where $x^i \in \{x_r, x_g\}$ represents real or generated images, and $y^i \in \{0, 1\}$ denotes their corresponding labels. 
Our method constructs the attribution manifold using samples from a single class, either only real images $x_r$ or only generated images $x_g$, enabling the model to learn class-consistent feature regularities.

The ACM aims to capture these regularities by learning a class-consistent transformation in feature space. 
It is implemented through an encoder–decoder mapping $\mathcal{D}\!\cdot\!\mathcal{E}:\mathbb{R}^D \!\rightarrow\! \mathbb{R}^D$, where $\mathcal{E}$ and $\mathcal{D}$ denote the encoder and decoder, respectively. 
The attribution mapping is formulated as:
\begin{equation}
    \mathcal{A}(f(x))=\mathrm{abs}\!\left(f(x)-\mathcal{D}(\mathcal{E}(f(x)))\right),
\end{equation}
where $\mathcal{A}(f(x))$ represents the attribution deviation, quantifying the feature-level discrepancy of a sample relative to the learned class manifold. By minimizing attribution deviations, the ACM learns the class-consistent structure and forms a compact feature manifold:
\begin{equation}
\mathcal{L}_{\mathcal{A}} = \mathbb{E}_{x_i \sim D_s^c}\|\mathcal{A}(f(x_i))\|_1,
\end{equation}
where $D_s^c \subset D_s$ denotes the single-class subset (either real or generated samples). Samples from the opposite domain naturally produce larger attribution deviations $\mathcal{A}(f(x_j))$ when projected into this space, providing measurable cues for detection.



\subsection{Attribution Deviations Discrimination}
After processing by the ACM, the original training set $D_s = \{(x^i, y^i)\}_{i=1}^{N}$ is transformed into an attribution deviation representation $D_{\mathrm{att}} = \{(\mathcal{A}(f(x^i)), y^i)\}_{i=1}^{N}$. 
Since $\mathcal{A}$ is trained using samples from a single class, features from the same class tend to produce smaller deviations, while those from the opposite class yield significantly larger and more irregular deviations. 
This difference results in distinguishable attribution patterns that implicitly encode the image’s authenticity.

A lightweight classifier $\mathcal{C}: \mathbb{R}^D \rightarrow [0,1]$ is employed to exploit these patterns for image authenticity discrimination. 
It takes the attribution deviation $\mathcal{A}(f(x^i))$ as input and outputs the probability that the image is generated. 
The classifier is optimized with a binary cross-entropy loss:
\begin{equation}
\mathcal{L}_\mathrm{classify} = \mathrm{BCE}(\mathcal{C}(\mathcal{A}(f(x^i))), y^i).
\end{equation}
This supervised discrimination reinforces the separability of attribution deviations and complements the class-consistent modeling learned by the ACM.

\begin{table*}[t]
    \centering
    \renewcommand{\arraystretch}{1.3}
    \setlength{\tabcolsep}{1mm}
    \resizebox{\linewidth}{!}{
    \begin{tabular}{ccccccccccccccccc}
    \toprule
    \toprule
    \multicolumn{1}{c}{\multirow{2}[4]{*}{Methods}} & 
    \multicolumn{6}{c}{Generative Adversarial Networks} & 
    \multicolumn{2}{c}{Others} & 
    \multicolumn{7}{c}{Diffusion Models} & 
    \textit{AP} \\
    
    \cmidrule(r){2-7} 
    \cmidrule(r){8-9} 
    \cmidrule(r){10-16} 
    \cmidrule(r){17-17}
    
    & \makecell{Pro-\\GAN} 
    & \makecell{Big-\\GAN} 
    & \makecell{Cycle-\\GAN} 
    & \makecell{Star-\\GAN} 
    & \makecell{Style-\\GAN} 
    & \makecell{Gau-\\GAN}    
    & SAN & \makecell{Deep-\\Fake}
    & ADM & GLIDE  & SD1.4 & SD1.5 & VQDM & Midj & Wukong & avg. \\
    
    \midrule
    CNNDet~\cite{wang2020cnn}& \textbf{100.00} & 84.52 &  93.46 & 98.15 & 99.54 & 89.49
    & 57.67 & 89.03
    & 71.05 & 66.09 & 56.88 & 57.25 & 61.90 & 55.89 & 52.85 
    & 75.58 \\
    
    
    
    LNP~\cite{liu2022detecting} & \underline{99.98} & 95.20 & 98.14 & \textbf{100.00} & 99.56 & 83.25 
    & 45.29 & 57.72 
    & \textbf{99.25} & \underline{96.45} & 89.77 & 89.30 & 95.54 & \textbf{94.39} & 91.16 
    & 89.00 \\
    
    LGrad~\cite{tan2023learning} & \textbf{100.00} & 90.76 & 94.02 & \underline{99.98} & \underline{99.81} & 79.29 
    & 45.10 & 71.71 
    & 71.84 & 75.96 & 70.91 & 71.73 & 70.23 & 71.43 & 66.51 
    & 78.62 \\

    DIRE~\cite{wang2023dire} & 69.76 & 51.92 & 64.76 & 67.98 & 71.40 & 88.18 
    & 54.89 & 67.90 
    & 44.99 & 58.11 & 47.46 & 47.06 & 85.18 & 55.55 & 51.25 & 61.76 \\
    
    UnivFD~\cite{ojha2023towards} & \textbf{100.00} & 99.23 & 99.76 & 99.01 & 97.88 & 99.98
    & 77.53 & 81.22 
    & 86.48 & 83.24 & 86.00 & 85.17 & 96.50 &  73.35 & 90.89
    & 90.45 \\

    CLIPing~\cite{khan2024clipping} & 99.85 & 91.27 & 94.03 & 99.06 & 95.24 & 89.53
    & 56.63 & 73.89
    & 77.26 & 77.94 & 58.29 & 57.86 & 80.00 & 52.44 & 60.65 & 77.60 \\

    FatFormer~\cite{liu2024forgery} & \textbf{100.00} & \textbf{99.98} & \textbf{100.00} & \textbf{100.00} & 99.75 & \textbf{100.00} 
    & 81.23 & \textbf{97.99}
    & 91.73 & 95.99 & 81.12 & 81.09 & \textbf{96.99} & 62.76 & 85.86 & 91.63 \\

    
    NPR~\cite{tan2024rethinking} &  \textbf{100.00} & 85.77 & 97.88 & \textbf{100.00} & \textbf{99.85} & 67.48
    & 76.76 & 91.88
    & 82.48 & 93.43 & 
    \textbf{94.03} & \textbf{94.17} & 71.88 & \underline{92.91} &  87.56 
    & 89.07 \\

    VIB-Net~\cite{zhang2025towards} & \textbf{100.00} & \underline{99.29} & \underline{99.80} & 99.72 & 98.79 & \underline{99.99}
    & \underline{91.62} & \underline{92.64} 
    & 87.88 & 88.53 & 87.24 & 86.98 & 96.51 & 75.68 & \underline{90.92} & \underline{93.04} \\
    
    \rowcolor{orange!10}
    Ours & \textbf{100.00} & 99.14 & 99.25 & 98.37 & 99.24 & 99.77
    & \textbf{93.63} & 81.24
    & \underline{93.21} & \textbf{97.26} & \underline{93.55} & \underline{93.45} & \underline{96.87} & 72.52 & \textbf{94.82} & \textbf{\textcolor{red}{94.15}} \\
    
    \bottomrule
    \bottomrule
    \end{tabular}
    }
    \caption{\textbf{The \textit{AP} values of various methods trained on GAN source images.} Bold values indicate the best performance, and underlined values indicate the second best.}
    \label{tab1:ap}%
\end{table*}

\begin{table*}[t]
    \centering
    \renewcommand{\arraystretch}{1.3}
    \setlength{\tabcolsep}{1mm}
    \resizebox{\linewidth}{!}{
    \begin{tabular}{ccccccccccccccccc}
    \toprule
    \toprule
    \multicolumn{1}{c}{\multirow{2}[4]{*}{Methods}} & 
    \multicolumn{6}{c}{Generative Adversarial Networks} & 
    \multicolumn{2}{c}{Others} & 
    \multicolumn{7}{c}{Diffusion Models} & 
    \textit{ACC} \\
    
    \cmidrule(r){2-7} 
    \cmidrule(r){8-9} 
    \cmidrule(r){10-16} 
    \cmidrule(r){17-17}
    
    & \makecell{Pro-\\GAN} 
    & \makecell{Big-\\GAN} 
    & \makecell{Cycle-\\GAN} 
    & \makecell{Star-\\GAN} 
    & \makecell{Style-\\GAN} 
    & \makecell{Gau-\\GAN}    
    & SAN & \makecell{Deep-\\Fake}
    & ADM & GLIDE  & SD1.4 & SD1.5 & VQDM & Midj & Wukong & avg. \\
    
    \midrule
    CNNDet~\cite{wang2020cnn} & \underline{99.99} & 70.08 &  85.16 & 91.65 & 85.73 & 78.94 
    & 50.91 & 53.49 
    & 58.73 & 54.96 & 51.57 & 51.78 & 53.64 & 52.61 & 50.23 
    & 65.96 \\
    
    
    
    LNP~\cite{liu2022detecting} & 99.09 & 88.05 & 91.63 & \textbf{100.00} & 95.89 & 79.70
    & 43.61 & 53.99 
    & \textbf{94.53} & 87.84 & 68.58 & 68.06 & 77.89 & \underline{79.92} & 73.89 
    & 80.18 \\
    
    LGrad~\cite{tan2023learning} & 99.83 & 82.90 & 85.28 & 99.60 & 94.66 & 72.49 
    & 44.47 & 56.42
    & 66.96 & 62.75 & 67.20 & 67.91 & 67.03 & 65.69 & 63.67 
    & 73.12 \\

    DIRE~\cite{wang2023dire} & 59.18 & 51.68 & 60.75 & 62.11 & 57.85 & 82.69 & 48.32 
    & 60.15 & 44.69
    & 59.14 & 47.53 & 47.02 & 73.83 & 53.13 & 50.09 & 57.21 \\
    
    UnivFD~\cite{ojha2023towards} & 99.85 & 95.08 & 98.18 & 93.75 & 85.37 & \underline{99.42}
    & 57.08 & 67.38 
    & 67.01 & 62.43 & 64.20 & 64.01 & 85.68 &  56.18 & 71.39 
    & 77.80 \\

    CLIPing~\cite{khan2024clipping} & 99.89 & 94.78 & 96.74 & 99.47 & 95.88 & 94.15
    & 60.27 & 76.48 
    & 81.78 & 82.69 & 61.27 & 60.78 & 84.77 & 53.96 & 63.94 & 80.46 \\

    FatFormer~\cite{liu2024forgery} & 99.89 & \textbf{99.50} & \textbf{99.36} & \underline{99.75} & \textbf{97.13} & \underline{99.42}
    & 68.04 & \textbf{93.27}
    & 78.45 & \underline{88.03} & 67.83 & 68.06 & \underline{86.88} & 56.09 & 73.06 & \underline{84.98} \\

    
    NPR~\cite{tan2024rethinking} & \textbf{100.00} & 79.50 & 84.94 & \textbf{100.00} & 93.98 & 68.09 
    & 69.63 & 72.19
    & 70.82 & 78.90 & \underline{81.46} & \underline{81.59} & 66.08 & \underline{79.50} &  \underline{78.09} 
    & 80.32 \\

    VIB-Net~\cite{zhang2025towards} & \underline{99.99} & 95.75 & \textbf{99.00} & 98.95 & 91.25 & \textbf{99.70} 
    & \underline{70.50} & \underline{83.20}
    & 71.45 & 69.40 & 71.55 & 70.00 & 86.65 & 61.25 & 75.90 & 82.97 \\

    \rowcolor{orange!10}
    Ours & \textbf{100.00} & \underline{96.28} & 98.56 & 97.42 & \underline{96.57} & \underline{99.42}
    & \textbf{91.78} & 77.11
    & \underline{89.00} & \textbf{94.95} & \textbf{89.96} & \textbf{90.33} & \textbf{92.39} & 70.48 & \textbf{91.14} & \textbf{\textcolor{red}{91.69}}\\
    \bottomrule
    \bottomrule
    \end{tabular}
    }
  \caption{\textbf{The \textit{ACC} values of various methods trained on GAN source images.} Symbols follow the same notation as Table~\ref {tab1:ap}.}
    \label{tab2:acc}%
    \vspace{-10pt}
\end{table*}

\section{Experiments}
In this section, we first introduce the experimental setup, and then report and discuss in detail the comparative analysis of our method with state-of-the-art methods.

\subsection{Experimental Setup}
\subsubsection{Training Dataset} 
For a fair comparison, we follow mainstream baselines and use the ProGAN training set from CNNDet~\cite{wang2020cnn}, and the SD1.4 training set from GenImage~\cite{zhu2023genimage}.

\subsubsection{Test Datasets} 
To evaluate the detection ability of our method in real-world scenarios, we test on a large collection of real and generated images from CNNDet and GenImage, covering 15 representative generative models. The CNNDet dataset includes three unconditional GANs (BigGAN~\cite{brock2018large}, ProGAN~\cite{karras2018progressive} and StyleGAN~\cite{karras2019style}), three conditional GANs (GanGAN~\cite{park2019semantic}, CycleGAN~\cite{zhu2017unpaired} and StarGAN~\cite{choi2018stargan}), one low-vision models (SAN~\cite{dai2019second}) and one face-swapping model (Deepfake~\cite{rao2016deep}). The GenImage dataset contains seven diffusion-based models: ADM~\cite{dhariwal2021diffusion}, Glide~\cite{nichol2022glide}, Stable Diffusion v1.4\&v1.5~\cite{rombach2022high}, VQDM~\cite{gu2022vector}, Midjourney~\cite{Midjourney}, and Wukong~\cite{wukong}. 

\subsubsection{Baselines} 
We compare our model to a set of state-of-the-art baselines that have released their training or inference code. These baselines include CNNDet (CVPR 20'), LNP (ECCV 22'), LGrad (CVPR 23'), DIRE (ICCV 23'), UnivFD (CVPR 23'), CLIPing (ICMR 24'), FatFormer (CVPR 24'), DRCT (ICML 24'), NPR (CVPR 24'), and VIB-Net (CVPR 25'). To ensure a fair comparison of generalization performance, all baselines are trained on the same datasets as our method.

\subsubsection{Evaluation Metrics} 
Following prior works, we use average precision (AP) and classification accuracy (ACC) as primary metrics. ACC is computed with a 0.5 threshold. To assess potential bias, we also report F1 scores, as well as real-image accuracy (Real ACC) and generated-image accuracy (Fake ACC) in some experiments.

\subsubsection{Implementation Details} 
Following baseline UnivFD, we use a pre-trained CLIP ViT-L/14 to extract image features with frozen weights. Both $\mathcal{A}$ and $\mathcal{C}$ are optimized using Adam with a learning rate of $2\times10^{-4}$ and a batch size of 256. All experiments are conducted on a single NVIDIA H100 GPU using PyTorch.

\subsection{Detection performance}

\subsubsection{Performance Evaluation Trained on DM Sources}
We compare the performance of our method with several SOTA detectors across 15 datasets, where all models are trained on the SD1.4 dataset. As shown in Fig.~\ref{fig4}, traditional data-driven methods such as CNNDet tend to overfit to specific distributions, resulting in poor cross-model performance. Contrastive or feature-based methods like CLIPing, UnivFD, and VIB-Net show improved generalization but still struggle on certain generators. NPR exhibits severe classification bias, as it tends to label most unseen samples as real, achieving near-100\% \textit{Real ACC} while \textit{Fake ACC} drops drastically (even close to 0\% on some datasets), indicating that it overfits to generator-specific artifacts. Diffusion-oriented methods (e.g., DIRE and DRCT) perform well on diffusion-based sources but fail to generalize to GAN-generated images. In contrast, our method achieves consistently superior performance and balanced scores between real and fake classes across all datasets, which captures intrinsic forensic regularities rather than generator-dependent artifacts, enabling strong generalization even to high-resolution and unseen diffusion models.

\begin{figure}[t]
    \centering
    \includegraphics[width=1\linewidth]{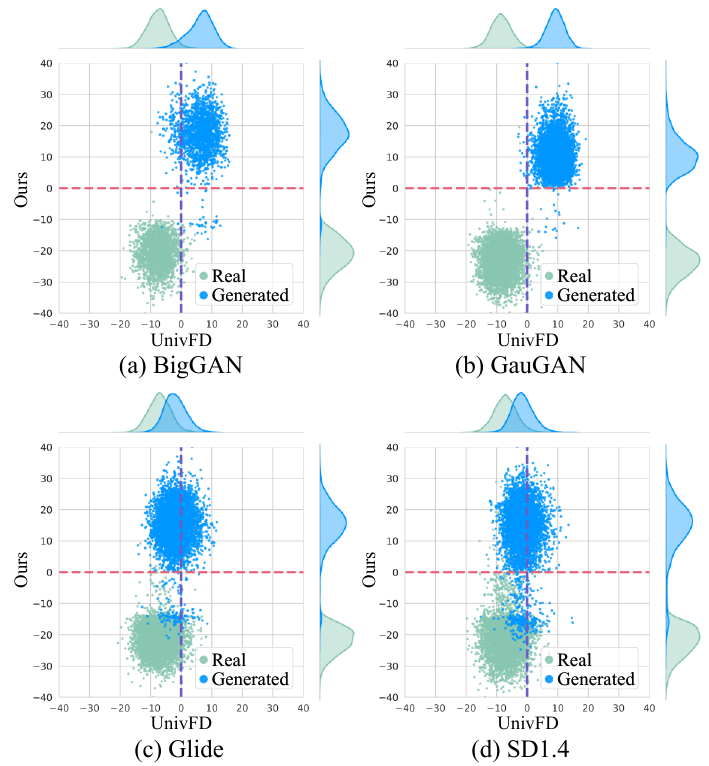}
    \caption{Scatter plots of per-sample logits. The X-axis and Y-axis represent the outputs of UnivFD and our method, respectively. Decision boundaries of UnivFD (purple) and ours (red) are shown. More results are in Appendix.}  
    \label{fig5}
\end{figure}

\subsubsection{Performance Valuation Trained on GAN Sources}
Table~\ref{tab1:ap} and Table~\ref{tab2:acc} present the \textit{AP} and \textit{ACC} of different methods trained on the ProGAN dataset and evaluated on 15 diverse datasets. 
Traditional data-driven and artifacts-based methods, such as CNNDet, LNP, and LGrad rely on modeling low-level traces, which makes them effective for training-related images but exhibits severe degradation on diffusion-based datasets. This indicates that these methods overfit to GAN-specific artifacts, limiting their cross-model generalization. 
DIRE exploits reconstruction priors from diffusion models, performs competitively on DM sources but fails to generalize to GAN architectures, indicating limited cross-model adaptability. Contrastive and feature-based methods like CLIPing, UnivFD, and VIB-Net, which leverage CLIP representations and fine-tuning strategies, show improved generalization compared to others, yet their performance remains unstable on challenging datasets like Midj and Wukong, where generative semantics differ significantly from training distributions. In contrast, our method achieves the best average results ($\textit{AP}=94.15\%$, $\textit{ACC}=91.69\%$), surpassing the second-best FatFormer by +6.71\% in \textit{ACC}. This advantage stems from modeling attribution consistency within a single-class manifold, which captures intrinsic forensic regularities rather than generator-dependent artifacts, enabling well generalization across both GAN and DM sources.

\subsection{Score Ensemble}
To further evaluate the generalization, we compare the logit distributions of the baseline UnivFD and our method across four representative datasets, as shown in Fig.~\ref{fig5}. UnivFD shows notable overlap near its decision boundary (purple dashed line), especially on challenging datasets such as Glide and SD1.4, where it frequently misclassifies images and tends to bias predictions toward real samples. In contrast, our method constructs a larger and more stable decision margin (red dashed line), achieving clear vertical separation and reducing classification ambiguity across all datasets. These results highlight the effectiveness of our attribution space in amplifying attribution deviations and achieving superior cross-model generalization.

\begin{table}[t]
    \centering
    \renewcommand{\arraystretch}{1.2}
    \resizebox{\linewidth}{!}{
    \begin{tabular}{c|cc|cc|cc|cc}
    \toprule
    
    \multirow{2}[2]{*}{Sources} 
    & \multicolumn{2}{c}{GANs} 
    & \multicolumn{2}{c}{DMs} 
    & \multicolumn{2}{c}{Others}
    & \multicolumn{2}{c}{avg.}
    \\
        
    \cmidrule(r){2-3} 
    \cmidrule(r){4-5} 
    \cmidrule(r){6-7} 
    \cmidrule(r){8-9} 

    & \textit{AP} & \textit{ACC} 
    & \textit{AP} & \textit{ACC} 
    & \textit{AP} & \textit{ACC} 
    & \textit{AP} & \textit{ACC} 
    \\
    \midrule

    Real (GAN) & 99.29 & 98.04 & 91.67 & 88.32 & 87.44 & 84.45 & 94.15 & 91.69 \\
    Real (DM) & 97.70 & 94.01 & 99.20 & 96.20 & 86.27 & 78.60 & 96.87 & 92.98 \\

    \midrule

    ProGAN & 99.18 & 98.09 & 90.96 & 87.78 & 89.44 & 84.06 & 94.05 & 91.41 \\
    ADM & 90.89 & 88.28 & 96.64 & 94.52 & 87.39 & 85.64 & 93.11 & 90.84 \\
    SD1.4 & 98.19 & 93.08 & 99.06 & 96.46 & 86.21 & 78.14 & 97.00 & 92.66 \\
    SD1.5 & 94.93 & 93.07 & 98.72 & 96.80 & 80.71 & 77.52 & 94.80 & 92.74 \\
    VQDM & 98.18 & 96.41 & 97.88 & 95.31 & 84.24 & 81.36 & 96.18 & 93.89 \\

    \bottomrule
    \end{tabular}
    }
    \caption{Performance comparison of models constructed with different attribution sources. Real (GAN) and Real (DM) indicate models built using real images from the ProGAN and SD1.4 datasets, respectively. Others use generated images only. More results are in Appendix.}
    \label{tab:3}
\end{table}

\subsection{Effectiveness of Various Attribution Sources}
Table~\ref{tab:3} examines the influence of different attribution sources on detection performance. Across all settings, our method maintains consistently high and balanced results on GANs, diffusion models, and other datasets, 
demonstrating its robustness and generalization regardless of the attribution source. 
Both real-image and generated-image constructions yield nearly identical \textit{AP} and \textit{ACC}, indicating that the proposed Attribution Space Modeling can effectively learn intrinsic class-consistent representations from either type of data. This further confirms that our framework does not rely on specific data distributions and can generalize well across diverse generative architectures.

\begin{figure}
    \centering
    \includegraphics[width=1\linewidth]{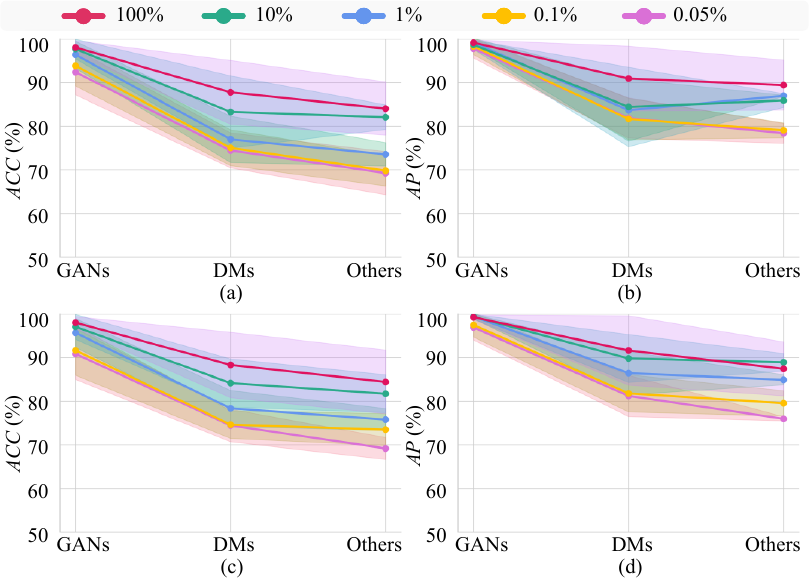}
    \caption{Performance under limited training data. We train the full pipeline using 0.05\%, 0.1\%, 1\%, 10\%, and 100\% of the ProGAN dataset (720K images in total). (a,b) Results on $\mathcal{M}_g$, and (c,d) results on $\mathcal{M}_r$.}  
    \label{fig6}
    \vspace{-3pt}
\end{figure}

\subsection{Effectiveness under Limited Training Data}
To evaluate the data efficiency of our method, we test its performance under different training data ratios. The training data is derived from ProGAN dataset, which contains 720k images. We randomly sample 0.05\% (360), 0.1\% (720), 1\% (7.2k), 10\% (72k), and 100\% (720k) of the data to train the full pipeline. As shown in Fig.~\ref{fig6}, our method maintains strong performance even with extremely limited training samples. When trained with only 0.05\% of the training data, the model already achieves competitive performance. As the training ratio increases, both metrics steadily improve and eventually saturate, suggesting that the attribution consistency module effectively captures stable class-specific regularities rather than relying on large-scale data.

A consistent trend is observed across both attribution spaces, $\mathcal{M}_g$ (Fig.~\ref{fig6} (a, b)) and $\mathcal{M}_r$ (Fig.~\ref{fig6} (c, d)), where the performance gap between the smallest and full datasets remains relatively small. This indicates that the proposed attribution space modeling can learn transferable and compact representations even from a small number of samples. Overall, these results confirm that the proposed approach achieves high data efficiency and strong generalization capability, making it suitable for practical scenarios where large-scale labeled data may not be available.

\subsection{Robustness to Perturbations}
\begin{figure}
    \centering
    \includegraphics[width=1\linewidth]{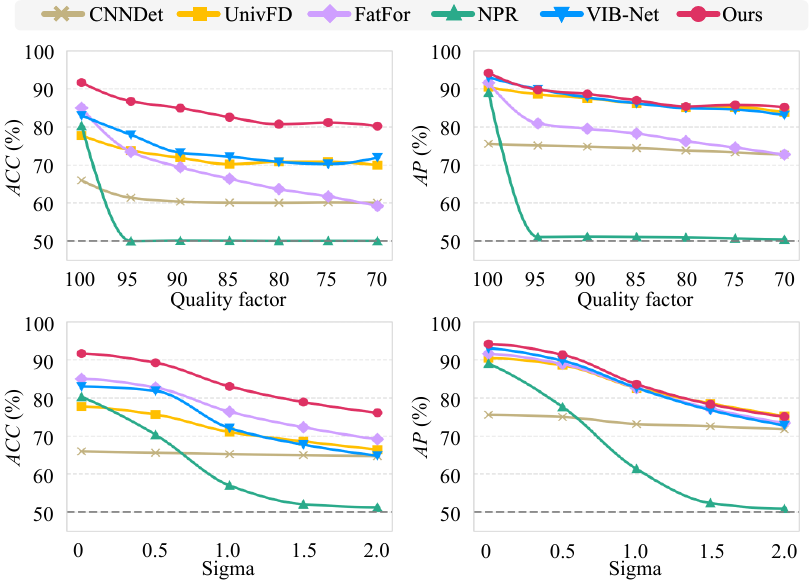}
    \caption{Robustness evaluation under perturbations with the top row showing JPEG compression and the bottom row showing Gaussian blur.}  
    \label{fig7}
    \vspace{-3pt}
\end{figure}

In real-world scenarios, images often undergo various post-processing operations. Therefore, beyond cross-dataset generalization, it is also essential to evaluate robustness against unknown perturbations. We assess detector performance under two common distortions: JPEG compression and Gaussian blur, as illustrated in Fig.~\ref{fig7}. 
As the perturbation intensity increases, most existing methods experience sharp performance degradation. NPR is particularly vulnerable, with accuracy dropping close to random under mild compression. CNNDet, UnivFD, and FatFormer also suffer significant declines, while VIB-Net shows moderate robustness. In contrast, our method consistently achieves the highest \textit{ACC} and \textit{AP} across all perturbation levels, demonstrating strong resistance to both compression and blurring effects. This robustness is from the attribution consistency modeling, which focuses on feature-level attribution deviations rather than post-processing-sensitive pixel artifacts.

\section{Conclusion}
We propose an attribution consistency framework for AI-generated image detection that overcomes the reliance of existing methods on data-dependent discriminative learning. By constructing an attribution space from a single-class training set, either real images or those generated by a single generative model, our method learns compact, class-consistent feature manifolds that enlarge the decision margins between real and generated images. Consequently, this perspective amplifies attribution deviations between learned and unseen distributions, leading to superior generalization and robustness across diverse generated images.

\section*{Acknowledgements}
This work was supported in part by the National Natural Science Foundation of China under Grant 62376046, Grant 62576060 and Grant 62536002, and in part by the Natural Science Foundation of Chongqing under Grant CSTB2023NSCQ-MSX0341, CSTB2025YITP-QCRCX0074.
{
    \small
    \bibliographystyle{ieeenat_fullname}
    \bibliography{main}

\begin{thebibliography}{42}
\providecommand{\natexlab}[1]{#1}
\providecommand{\url}[1]{\texttt{#1}}
\expandafter\ifx\csname urlstyle\endcsname\relax
  \providecommand{\doi}[1]{doi: #1}\else
  \providecommand{\doi}{doi: \begingroup \urlstyle{rm}\Url}\fi

\bibitem[Brock et~al.(2018)Brock, Donahue, and Simonyan]{brock2018large}
Andrew Brock, Jeff Donahue, and Karen Simonyan.
\newblock Large scale gan training for high fidelity natural image synthesis.
\newblock \emph{arXiv preprint arXiv:1809.11096}, 2018.

\bibitem[Cazenavette et~al.(2024)Cazenavette, Sud, Leung, and Usman]{cazenavette2024fakeinversion}
George Cazenavette, Avneesh Sud, Thomas Leung, and Ben Usman.
\newblock Fakeinversion: Learning to detect images from unseen text-to-image models by inverting stable diffusion.
\newblock In \emph{Proceedings of the IEEE/CVF Conference on Computer Vision and Pattern Recognition}, pages 10759--10769, 2024.

\bibitem[Chandrasegaran et~al.(2021)Chandrasegaran, Tran, and Cheung]{chandrasegaran2021closer}
Keshigeyan Chandrasegaran, Ngoc-Trung Tran, and Ngai-Man Cheung.
\newblock A closer look at fourier spectrum discrepancies for cnn-generated images detection.
\newblock In \emph{Proceedings of the IEEE/CVF conference on computer vision and pattern recognition}, pages 7200--7209, 2021.

\bibitem[Chen et~al.(2024)Chen, Zeng, Yang, and Yang]{chen2024drct}
Baoying Chen, Jishen Zeng, Jianquan Yang, and Rui Yang.
\newblock Drct: Diffusion reconstruction contrastive training towards universal detection of diffusion generated images.
\newblock In \emph{Forty-first International Conference on Machine Learning}, 2024.

\bibitem[Choi et~al.(2018)Choi, Choi, Kim, Ha, Kim, and Choo]{choi2018stargan}
Yunjey Choi, Minje Choi, Munyoung Kim, Jung-Woo Ha, Sunghun Kim, and Jaegul Choo.
\newblock Stargan: Unified generative adversarial networks for multi-domain image-to-image translation.
\newblock In \emph{Proceedings of the IEEE conference on computer vision and pattern recognition}, pages 8789--8797, 2018.

\bibitem[Corvi et~al.(2023)Corvi, Cozzolino, Zingarini, Poggi, Nagano, and Verdoliva]{corvi2023detection}
Riccardo Corvi, Davide Cozzolino, Giada Zingarini, Giovanni Poggi, Koki Nagano, and Luisa Verdoliva.
\newblock On the detection of synthetic images generated by diffusion models.
\newblock In \emph{ICASSP 2023-2023 IEEE International Conference on Acoustics, Speech and Signal Processing (ICASSP)}, pages 1--5. IEEE, 2023.

\bibitem[Cozzolino et~al.(2024)Cozzolino, Poggi, Corvi, Nie{\ss}ner, and Verdoliva]{cozzolino2024raising}
Davide Cozzolino, Giovanni Poggi, Riccardo Corvi, Matthias Nie{\ss}ner, and Luisa Verdoliva.
\newblock Raising the bar of ai-generated image detection with clip.
\newblock In \emph{Proceedings of the IEEE/CVF Conference on Computer Vision and Pattern Recognition}, pages 4356--4366, 2024.

\bibitem[Dai et~al.(2019)Dai, Cai, Zhang, Xia, and Zhang]{dai2019second}
Tao Dai, Jianrui Cai, Yongbing Zhang, Shu-Tao Xia, and Lei Zhang.
\newblock Second-order attention network for single image super-resolution.
\newblock In \emph{Proceedings of the IEEE/CVF conference on computer vision and pattern recognition}, pages 11065--11074, 2019.

\bibitem[Dhariwal and Nichol(2021)]{dhariwal2021diffusion}
Prafulla Dhariwal and Alexander Nichol.
\newblock Diffusion models beat gans on image synthesis.
\newblock \emph{Advances in neural information processing systems}, 34:\penalty0 8780--8794, 2021.

\bibitem[Esser et~al.(2021)Esser, Rombach, and Ommer]{esser2021taming}
Patrick Esser, Robin Rombach, and Bjorn Ommer.
\newblock Taming transformers for high-resolution image synthesis.
\newblock In \emph{Proceedings of the IEEE/CVF conference on computer vision and pattern recognition}, pages 12873--12883, 2021.

\bibitem[Frank et~al.(2020)Frank, Eisenhofer, Sch{\"o}nherr, Fischer, Kolossa, and Holz]{frank2020leveraging}
Joel Frank, Thorsten Eisenhofer, Lea Sch{\"o}nherr, Asja Fischer, Dorothea Kolossa, and Thorsten Holz.
\newblock Leveraging frequency analysis for deep fake image recognition.
\newblock In \emph{International conference on machine learning}, pages 3247--3258. PMLR, 2020.

\bibitem[Gu et~al.(2022)Gu, Chen, Bao, Wen, Zhang, Chen, Yuan, and Guo]{gu2022vector}
Shuyang Gu, Dong Chen, Jianmin Bao, Fang Wen, Bo Zhang, Dongdong Chen, Lu Yuan, and Baining Guo.
\newblock Vector quantized diffusion model for text-to-image synthesis.
\newblock In \emph{Proceedings of the IEEE/CVF conference on computer vision and pattern recognition}, pages 10696--10706, 2022.

\bibitem[Guillaro et~al.(2025)Guillaro, Zingarini, Usman, Sud, Cozzolino, and Verdoliva]{guillaro2025bias}
Fabrizio Guillaro, Giada Zingarini, Ben Usman, Avneesh Sud, Davide Cozzolino, and Luisa Verdoliva.
\newblock A bias-free training paradigm for more general ai-generated image detection.
\newblock In \emph{Proceedings of the Computer Vision and Pattern Recognition Conference}, pages 18685--18694, 2025.

\bibitem[Gye et~al.(2025)Gye, Ko, Shon, Kwon, and Kim]{gye2025reducing}
Seoyeon Gye, Junwon Ko, Hyounguk Shon, Minchan Kwon, and Junmo Kim.
\newblock Reducing the content bias for ai-generated image detection.
\newblock In \emph{2025 IEEE/CVF Winter Conference on Applications of Computer Vision (WACV)}, pages 399--408. IEEE, 2025.

\bibitem[He et~al.(2026)He, Zhang, Qin, Liu, Bi, and Xiao]{he2026diversity}
Qinghui He, Haifeng Zhang, Qiao Qin, Bo Liu, Xiuli Bi, and Bin Xiao.
\newblock Diversity over uniformity: Rethinking representation in generated image detection.
\newblock \emph{arXiv preprint arXiv:2603.00717}, 2026.

\bibitem[Karras et~al.(2017)Karras, Aila, Laine, and Lehtinen]{karras2018progressive}
Tero Karras, Timo Aila, Samuli Laine, and Jaakko Lehtinen.
\newblock Progressive growing of gans for improved quality, stability, and variation.
\newblock \emph{arXiv preprint arXiv:1710.10196}, 2017.

\bibitem[Karras et~al.(2019)Karras, Laine, and Aila]{karras2019style}
Tero Karras, Samuli Laine, and Timo Aila.
\newblock A style-based generator architecture for generative adversarial networks.
\newblock In \emph{Proceedings of the IEEE/CVF conference on computer vision and pattern recognition}, pages 4401--4410, 2019.

\bibitem[Khan and Dang-Nguyen(2024)]{khan2024clipping}
Sohail~Ahmed Khan and Duc-Tien Dang-Nguyen.
\newblock Clipping the deception: Adapting vision-language models for universal deepfake detection.
\newblock In \emph{Proceedings of the 2024 International Conference on Multimedia Retrieval}, pages 1006--1015, 2024.

\bibitem[Liu et~al.(2022)Liu, Yang, Bi, Xiao, Li, and Gao]{liu2022detecting}
Bo Liu, Fan Yang, Xiuli Bi, Bin Xiao, Weisheng Li, and Xinbo Gao.
\newblock Detecting generated images by real images.
\newblock In \emph{European conference on computer vision}, pages 95--110. Springer, 2022.

\bibitem[Liu et~al.(2024)Liu, Tan, Tan, Wei, Wang, and Zhao]{liu2024forgery}
Huan Liu, Zichang Tan, Chuangchuang Tan, Yunchao Wei, Jingdong Wang, and Yao Zhao.
\newblock Forgery-aware adaptive transformer for generalizable synthetic image detection.
\newblock In \emph{Proceedings of the IEEE/CVF Conference on Computer Vision and Pattern Recognition}, pages 10770--10780, 2024.

\bibitem[Luo et~al.(2024)Luo, Du, Yan, and Ding]{luo2024lare}
Yunpeng Luo, Junlong Du, Ke Yan, and Shouhong Ding.
\newblock Lare\^{} 2: Latent reconstruction error based method for diffusion-generated image detection.
\newblock In \emph{Proceedings of the IEEE/CVF Conference on Computer Vision and Pattern Recognition}, pages 17006--17015, 2024.

\bibitem[Midjourney()]{Midjourney}
Midjourney.
\newblock Midjourney.
\newblock \url{https://www.midjourney.com}.
\newblock 2024.

\bibitem[MindSpore()]{wukong}
Huawei MindSpore.
\newblock Wukong.
\newblock \url{https://xihe.mindspore.cn/modelzoo/wukong/introduce}.
\newblock 2024.

\bibitem[Nichol et~al.(2022)Nichol, Dhariwal, Ramesh, Shyam, Mishkin, Mcgrew, Sutskever, and Chen]{nichol2022glide}
Alexander~Quinn Nichol, Prafulla Dhariwal, Aditya Ramesh, Pranav Shyam, Pamela Mishkin, Bob Mcgrew, Ilya Sutskever, and Mark Chen.
\newblock Glide: Towards photorealistic image generation and editing with text-guided diffusion models.
\newblock In \emph{International Conference on Machine Learning}, pages 16784--16804. PMLR, 2022.

\bibitem[Ojha et~al.(2023)Ojha, Li, and Lee]{ojha2023towards}
Utkarsh Ojha, Yuheng Li, and Yong~Jae Lee.
\newblock Towards universal fake image detectors that generalize across generative models.
\newblock In \emph{Proceedings of the IEEE/CVF conference on computer vision and pattern recognition}, pages 24480--24489, 2023.

\bibitem[Park et~al.(2019)Park, Liu, Wang, and Zhu]{park2019semantic}
Taesung Park, Ming-Yu Liu, Ting-Chun Wang, and Jun-Yan Zhu.
\newblock Semantic image synthesis with spatially-adaptive normalization.
\newblock In \emph{Proceedings of the IEEE/CVF conference on computer vision and pattern recognition}, pages 2337--2346, 2019.

\bibitem[Radford et~al.(2021)Radford, Kim, Hallacy, Ramesh, Goh, Agarwal, Sastry, Askell, Mishkin, Clark, et~al.]{radford2021learning}
Alec Radford, Jong~Wook Kim, Chris Hallacy, Aditya Ramesh, Gabriel Goh, Sandhini Agarwal, Girish Sastry, Amanda Askell, Pamela Mishkin, Jack Clark, et~al.
\newblock Learning transferable visual models from natural language supervision.
\newblock In \emph{International conference on machine learning}, pages 8748--8763. PmLR, 2021.

\bibitem[Ramesh et~al.(2022)Ramesh, Dhariwal, Nichol, Chu, and Chen]{ramesh2022hierarchical}
Aditya Ramesh, Prafulla Dhariwal, Alex Nichol, Casey Chu, and Mark Chen.
\newblock Hierarchical text-conditional image generation with clip latents.
\newblock \emph{arXiv preprint arXiv:2204.06125}, 1\penalty0 (2):\penalty0 3, 2022.

\bibitem[Rao and Ni(2016)]{rao2016deep}
Yuan Rao and Jiangqun Ni.
\newblock A deep learning approach to detection of splicing and copy-move forgeries in images.
\newblock In \emph{2016 IEEE international workshop on information forensics and security (WIFS)}, pages 1--6. IEEE, 2016.

\bibitem[Razavi et~al.(2019)Razavi, Van~den Oord, and Vinyals]{razavi2019generating}
Ali Razavi, Aaron Van~den Oord, and Oriol Vinyals.
\newblock Generating diverse high-fidelity images with vq-vae-2.
\newblock \emph{Advances in neural information processing systems}, 32, 2019.

\bibitem[Ricker et~al.(2024)Ricker, Lukovnikov, and Fischer]{ricker2024aeroblade}
Jonas Ricker, Denis Lukovnikov, and Asja Fischer.
\newblock Aeroblade: Training-free detection of latent diffusion images using autoencoder reconstruction error.
\newblock In \emph{Proceedings of the IEEE/CVF Conference on Computer Vision and Pattern Recognition}, pages 9130--9140, 2024.

\bibitem[Rombach et~al.(2022)Rombach, Blattmann, Lorenz, Esser, and Ommer]{rombach2022high}
Robin Rombach, Andreas Blattmann, Dominik Lorenz, Patrick Esser, and Bj{\"o}rn Ommer.
\newblock High-resolution image synthesis with latent diffusion models.
\newblock In \emph{Proceedings of the IEEE/CVF conference on computer vision and pattern recognition}, pages 10684--10695, 2022.

\bibitem[Tan et~al.(2023)Tan, Zhao, Wei, Gu, and Wei]{tan2023learning}
Chuangchuang Tan, Yao Zhao, Shikui Wei, Guanghua Gu, and Yunchao Wei.
\newblock Learning on gradients: Generalized artifacts representation for gan-generated images detection.
\newblock In \emph{Proceedings of the IEEE/CVF Conference on Computer Vision and Pattern Recognition}, pages 12105--12114, 2023.

\bibitem[Tan et~al.(2024)Tan, Zhao, Wei, Gu, Liu, and Wei]{tan2024rethinking}
Chuangchuang Tan, Yao Zhao, Shikui Wei, Guanghua Gu, Ping Liu, and Yunchao Wei.
\newblock Rethinking the up-sampling operations in cnn-based generative network for generalizable deepfake detection.
\newblock In \emph{Proceedings of the IEEE/CVF conference on computer vision and pattern recognition}, pages 28130--28139, 2024.

\bibitem[Tan et~al.(2025)Tan, Tao, Liu, Gu, Wu, Zhao, and Wei]{tan2025c2p}
Chuangchuang Tan, Renshuai Tao, Huan Liu, Guanghua Gu, Baoyuan Wu, Yao Zhao, and Yunchao Wei.
\newblock C2p-clip: Injecting category common prompt in clip to enhance generalization in deepfake detection.
\newblock In \emph{Proceedings of the AAAI Conference on Artificial Intelligence}, pages 7184--7192, 2025.

\bibitem[Wang et~al.(2020)Wang, Wang, Zhang, Owens, and Efros]{wang2020cnn}
Sheng-Yu Wang, Oliver Wang, Richard Zhang, Andrew Owens, and Alexei~A Efros.
\newblock Cnn-generated images are surprisingly easy to spot... for now.
\newblock In \emph{Proceedings of the IEEE/CVF conference on computer vision and pattern recognition}, pages 8695--8704, 2020.

\bibitem[Wang et~al.(2023)Wang, Bao, Zhou, Wang, Hu, Chen, and Li]{wang2023dire}
Zhendong Wang, Jianmin Bao, Wengang Zhou, Weilun Wang, Hezhen Hu, Hong Chen, and Houqiang Li.
\newblock Dire for diffusion-generated image detection.
\newblock In \emph{Proceedings of the IEEE/CVF International Conference on Computer Vision}, pages 22445--22455, 2023.

\bibitem[Wu et~al.(2025)Wu, Zhou, and Zhang]{wu2025generalizable}
Haiwei Wu, Jiantao Zhou, and Shile Zhang.
\newblock Generalizable synthetic image detection via language-guided contrastive learning.
\newblock \emph{IEEE Transactions on Artificial Intelligence}, 2025.

\bibitem[Yan et~al.(2025)Yan, Wang, Jin, Zhang, Liu, Chen, Yao, Ding, Wu, and Yuan]{yan2024effort}
Zhiyuan Yan, Jiangming Wang, Peng Jin, Ke-Yue Zhang, Chengchun Liu, Shen Chen, Taiping Yao, Shouhong Ding, Baoyuan Wu, and Li Yuan.
\newblock Orthogonal subspace decomposition for generalizable ai-generated image detection.
\newblock In \emph{International Conference on Machine Learning}, pages 70268--70288. PMLR, 2025.

\bibitem[Zhang et~al.(2025)Zhang, He, Bi, Li, Liu, and Xiao]{zhang2025towards}
Haifeng Zhang, Qinghui He, Xiuli Bi, Weisheng Li, Bo Liu, and Bin Xiao.
\newblock Towards universal ai-generated image detection by variational information bottleneck network.
\newblock In \emph{Proceedings of the Computer Vision and Pattern Recognition Conference}, pages 23828--23837, 2025.

\bibitem[Zhu et~al.(2017)Zhu, Park, Isola, and Efros]{zhu2017unpaired}
Jun-Yan Zhu, Taesung Park, Phillip Isola, and Alexei~A Efros.
\newblock Unpaired image-to-image translation using cycle-consistent adversarial networks.
\newblock In \emph{Proceedings of the IEEE international conference on computer vision}, pages 2223--2232, 2017.

\bibitem[Zhu et~al.(2023)Zhu, Chen, Yan, Huang, Lin, Li, Tu, Hu, Hu, and Wang]{zhu2023genimage}
Mingjian Zhu, Hanting Chen, Qiangyu Yan, Xudong Huang, Guanyu Lin, Wei Li, Zhijun Tu, Hailin Hu, Jie Hu, and Yunhe Wang.
\newblock Genimage: A million-scale benchmark for detecting ai-generated image.
\newblock \emph{Advances in neural information processing systems}, 36:\penalty0 77771--77782, 2023.

\end{thebibliography}
}


\end{document}